%% file: eccv2022submission.tex
\documentclass[runningheads]{llncs}
\usepackage{graphicx}
\usepackage{tikz}
\usepackage{comment}
\usepackage{amsmath,amssymb} % define this before the line numbering.
\usepackage{color}
\usepackage[width=122mm,left=12mm,paperwidth=146mm,height=193mm,top=12mm,paperheight=217mm]{geometry}
\usepackage{graphicx}
\usepackage{booktabs}
\usepackage[pagebackref,breaklinks,colorlinks]{hyperref}

\usepackage[accsupp]{axessibility}  % Improves PDF readability for those with disabilities.

\usepackage[width=122mm,left=12mm,paperwidth=146mm,height=193mm,top=12mm,paperheight=217mm]{geometry}

\begin{document}
\newcommand{\todo}[1]{\textcolor{red}{(TODO: #1)}}
% \linenumbers
\pagestyle{headings}
\mainmatter
\def\ECCVSubNumber{2152}  % Insert your submission number here

\title{Fine-Grained Egocentric Hand-Object Segmentation: Dataset, Model, and Applications} % Replace with your title

%\begin{comment}
\titlerunning{ECCV-22 submission ID \ECCVSubNumber} 
\authorrunning{ECCV-22 submission ID \ECCVSubNumber} 
\author{Anonymous ECCV submission}
\institute{Paper ID \ECCVSubNumber}
%\end{comment}
%******************

% \begin{comment}
\titlerunning{Egocentric Hand-Object Segmentation}
\author{Lingzhi Zhang*\inst{1} \and
Shenghao Zhou*\inst{1} \and Simon Stent\inst{2} \and Jianbo Shi\inst{1}}
\authorrunning{Zhang and Zhou et al.}
\institute{University of Pennsylvania \and
Toyota Research Institute \\
}
% \end{comment}
%******************
\maketitle
\footnotetext[1]{* indicates equal contribution}

\begin{abstract}

Egocentric videos offer fine-grained information for high-fidelity modeling of human behaviors. Hands and interacting objects are one crucial aspect of understanding a viewer's behaviors and intentions. We provide a labeled dataset consisting of 11,243 egocentric images with per-pixel segmentation labels of hands and objects being interacted with during a diverse array of daily activities. Our dataset is the first to label detailed hand-object contact boundaries. We introduce a context-aware compositional data augmentation technique to adapt to out-of-distribution YouTube egocentric video. We show that our robust hand-object segmentation model and dataset can serve as a foundational tool to boost or enable several downstream vision applications, including hand state classification, video activity recognition, 3D mesh reconstruction of hand-object interactions, and video inpainting of hand-object foregrounds in egocentric videos. Dataset and code are available at: {\href{https://github.com/owenzlz/EgoHOS}{https://github.com/owenzlz/EgoHOS}}

% Egocentric videos are an interesting source to analyze and understand human behaviors in a fine-level. Hands and interacting objects are, of course, the key component to reflect egocentric viewer's behaviors and intentions. In this work, we provide a labeled dataset consisting of 11,243 egocentric images with per-pixel segmentation labels of hands and the interacting objects in diverse daily activities. Our proposed dataset enable the segmentation model to generalize well in the out-of-the-distribution YouTube egocentric videos, significantly outperforming the previous labeled datasets. Since our dataset is the first to label interacting object mask, we also benchmark the hand-object segmentation with a proposed causal segmentation model, which is proven to be superior than the baselines. We also introduce a context-aware compositional data augmentation technique, which effectively boosts the object segmentation mIoU by approximately 2$\times$. Finally, we show that our robust hand-object segmentation model can serve as a foundation tool to boost or enable several vision applications, such as hand state classification, video activity recognition, 3D mesh reconstruction of hand-object interaction, and seeing through the hand with video inpainting in egocentric videos. All of our data and code will be released to the public. 


\end{abstract}

\section{Introduction}
\label{sec:introduction}

% Why study hand and interacting object? 

%  ``what kinds of veggies/meats does the person cut on the cupboard, and to what extent?", ``whether the person fry the egg on one-side or two-side in a stir fry pan?"

Watching someone cooking from a third-person view, we can answer questions such as ``what food is the person making?", or ``what cooking technique is the person using?" First-person egocentric video, on the other hand, can often show much more detailed information of human behaviors, such as ``what finger poses are needed to cut a steak into slices?", ``what are the procedures to construct a IKEA table with all the pieces and screws?" Thus, egocentric videos are an essential source of information to study and understand how humans interact with the world at a fine level. In these videos, egocentric viewer's hands and interacting objects are incredibly informative visual cues to understand human behaviors. However, existing tools for extracting these cues are limited, due to lack of robustness in the wild or coarse hand-object representation. Our goal is to create data labels and data argumentation tools for a robust fine-grained egocentric hand-object segmentation system that can generalize in the wild. Utilizing the fine-level interaction segmentation, we show how to construct a high-fidelity model that can serve as a foundation for understanding and modeling human hand-object behaviors. 

%Imagine we are watching someone cooking in a third-person video, we would likely read information about "what food the person is making?", "when does the person finishes cooking or start cleaning?", and so on. The egocentric video, on the other hand, could often show much more detailed information of human behaviors, such as "what kinds of veggies/meats does the person cut on the cupboard, and to what extent?", "whether the person fry the egg on one-side or two-side in a stir fry pan?" and so on. Thus, egocentric videos are an interesting source of information to study and understand how human interact with the world at a fine-level. Inside these videos, egocentric viewer's hands and interacting objects are, of course, especially informative visual cues to understand human behaviors. In this work, we aim to build a robust egocentric hand-object segmentation system that can generalize in the wild, where the model can be served as a foundation for understanding and modeling human behaviors. 

% dataset
The first and foremost factor in building a robust egocentric hand-object segmentation model is a good-quality labeled dataset. Previous works \cite{bambach2015lending,li2018eye,urooj2018analysis} have constructed hand segmentation datasets for egocentric videos. However, the collected data is mostly restricted to in-lab settings or to limited scenes, and lack labels for interacting objects. More recently, 100-DOH \cite{shan2020understanding} made a great effort to label large-scale hand and object interactions in the wild, but the labels for hands and objects are at the bounding box level. To bridge the gap and further advance fine-level understanding of hand-object interactions, we propose a new dataset of 11,243 images with per-pixel segmentation labels. A major characteristic is that our dataset contains very diverse hand-object interaction activities and scenarios, where frames are sparsely sampled from nearly 1,000 videos in Ego4D \cite{grauman2021ego4d}, EPIC-KITCHEN \cite{damen2018scaling}, THU-READ \cite{tang2017action}, and from our own collected GoPro videos. In addition, we also provide fine-grained labels of whether an object is interacted with by the left hand, right hand, or both hands and whether it is being interacted with directly (in touch) or indirectly. 

%The first and foremost factor to build a robust egocentric hand-object segmentation model is a good-quality dataset. Previous works have constructed hand segmentation datasets for egocentric videos, but the collected data are mostly restricted in the lab settings or with limited scenes and without any labels for interacting objects. More recently, 100-DOH \cite{shan2020understanding} has made a great effort to label large-scale hand and object interactions in the wild, but the labels for hands and objects are at the bounding box level. To bridge the gap and further advance fine-level understanding of hand-object interaction, we propose a new dataset of 11,243 images with per-pixel segmentation labels. A major characteristic is that our dataset contains very diverse hand-object interaction activities and scenarios, where the frames are sparsely sampled from nearly 1,000 videos in Ego4D \cite{grauman2021ego4d}, EPIC-KITCHEN \cite{damen2018scaling}, THU-READ \cite{tang2017action}, and our own collected GoPro videos. In addition, we also provide fine-grained labels of whether an object is interacted by left hand, right hand, or both hand, and whether it is in a direct and indirect interaction. 
To serve as an out-of-distribution test set for evaluating in-the-wild performance, we sparsely sampled and labeled 500 additional frames from 30 egocentric videos on YouTube. With our new segmentation dataset, we boost the hand segmentation performance significantly compared with the previous datasets \cite{bambach2015lending,li2018eye,urooj2018analysis}. Our dataset is the first to label interacting hand-object contact boundaries in egocentric videos. We show this label can improve the detection and segmentation of interaction objects. No matter how diverse our dataset is, we will inevitably encounter new domains with very different illumination, objects, and background clutter. We propose a context-aware data augmentation technique that adaptively composites hand-object pairs into diverse but plausible backgrounds. Our experiments show that our method is effective for out-of-domain adaptation.

%ith our new segmentation dataset, we boost the in-the-wild hand segmentation performance by a significant margin compared to the previous datasets. To evaluate the in-the-wild performance, we further collected and labeled 500 frames from 30 Youtube egocentric videos as our out-of-distribution test set. Our dataset is the first to label interacting object masks in egocentric videos, where object segmentation is a much harder task than hand segmentation due to the object diversity. To tackle this problem, we introduce dense contact boundary to explicitly model hand-object relationship, which effectively improves segmentation performance. In addition, we proposed a context-aware data augmentation technique that composite hand-object into diverse but plausible backgrounds. According to our ablation study, our proposed data augmentation technique is also proven to be very useful to boost object segmentation. 

% Since previous works have never provided segmentation labels for interacting objects, we also proposed a model and provided the hand-object segmentation benchmark performance, and compared with several baselines.  

\input{figs/overview}

% model and downstream applications: hand state/activity recognition, mesh reconstruction of hand-object interaction, seeing through the hands

We view our hand-object segmentation model as a foundation for boosting or enabling many vision applications, of which we demonstrate three, as shown in Fig. \ref{fig:overview}. First, we show that recognition tasks can get consistent performance improvement by simply adding reliably segmented hand or object masks as inputs. We experiment with a low-level recognition task to classify the left/right-hand state and a high-level recognition task to understand egocentric video activities by predicting verbs and nouns. Another useful but challenging application is reconstructing hand-object interaction in 3D mesh, which relies on the 2D hand-object masks during optimization. In this application, we integrate our hand-object segmentation model into the mesh reconstruction pipeline \cite{hasson2021towards}, and show improvements and generalization for mesh reconstruction of hand-object, compared to its original hand-object segmentation pipeline pretrained on COCO \cite{lin2014microsoft}. Finally, we show an interesting application by combining our accurate per-frame hand segmentation and video inpainting \cite{gao2020flow} to see through hands in egocentric videos, which could help scene understanding models that have not been trained with hands in the foreground. More details of each of these applications are discussed in Section \ref{sec:applications}. 

We summarize the contributions of this work as follows: 1) We propose a dataset of 11,243 images with fine-grained per-pixel labels of hand and interacting objects, including interacting object masks, enabling hand segmentation models to generalize much better than previous datasets. 2) We introduce the notion of a dense contact boundary to explicitly model the relationship between hands and interacting objects, which we show helps to improve segmentation performance. 3) We propose a context-aware compositional data augmentation technique, which effectively boosts object segmentation. 4) We demonstrate that our system can serve as a reliable foundational tool to boost or enable many vision applications, such as hand state classification, video activity recognition, 3D reconstruction of hand-object interaction, and seeing through hands in egocentric videos. We will release our dataset, code, and checkpoints to the public for future research.

\vspace{-5 pt}
\section{Related Work}
\vspace{-5 pt}
\label{sec:related_work}

\subsection{Hand Segmentation}

Prior to deep learning, several works have attempted to solve the hand segmentation task. Jedynak et al. \cite{jones2002statistical} used a color statistics--based approach to separate the skin region and the background. Ren and Gu \cite{ren2010figure} proposed a bottom-up motion-based approach to segment hand and object using the different motion patterns between hands and background in egocentric videos. Following up with \cite{ren2010figure}, Fathi et al. \cite{fathi2011learning} further separates hand and interacting object from the whole foreground by assuming a color histogram prior over hand super-pixels, and uses graph-cut to segment hands and objects. Li and Kitani \cite{li2013model,li2013pixel} first addressed the hand segmentation problem under various illuminations, and proposed to adaptively select a model that works the best under different illumination scenarios during inference time. Zhu et al. \cite{zhu2014pixel} proposed a novel approach by estimating a probability shape mask for a pixel using shape-aware structured forests. Beyond the egocentric viewer, Lee et al. \cite{lee2014hand} studied the problem of hand disambiguation of multiple people in egocentric videos by modeling the spatial, temporal, and appearance coherency constraints of moving hands.

More recently, many works \cite{bambach2015lending,urooj2018analysis,cai2020generalizing,li2015delving,lin2020ego2hands,lin2021two,shan2020understanding,shan2021cohesiv,shilkrot2019workinghands,kim2020first,narasimhaswamy2020detecting} have applied deep networks for hand or object segmentation. Bambach et al. \cite{bambach2015lending} introduced a dataset that contains 48 egocentric video clips for people interacting with others in real environments with over 15,000 labeled hand instances. The authors also proposed CNNs to first detect hand bounding boxes and then use GrabCut \cite{rother2004grabcut} to segment the hands. Following up on the same dataset, Urooj and Borji \cite{urooj2018analysis} used the RefineNet-ResNet101 \cite{lin2017refinenet} to achieve the state-of-the-art hand segmentation performance at the time. To alleviate the generalization issue, Cai et al. \cite{cai2020generalizing} proposed the use of a Bayesian CNN to predict the model uncertainty and leveraged the common information of hand shapes to better adapt to an unseen domain. There are also some other dataset efforts regarding hand segmentation. Li et al. \cite{li2015delving} proposed the Georgia Tech Egocentric Activity Datasets (GTEA), which includes 625 frames with two-hand labeling and 38 frames with binary labeling. Later, Li et al. \cite{li2018eye} extended the dataset (EGTEA) with 1,046 frames with two-hand labels and 12,799 frames with binary masks. Lin et al. \cite{lin2020ego2hands,lin2021two} also explored artificially composited hands with various backgrounds to scale up a large-scale synthetic dataset. Urooj et al. \cite{urooj2018analysis} recognized the constrained environment as one big limitation of existing datasets and collected an in-the-wild dataset by sampling frames from YouTube videos (EYTH). Though it is more diverse, it is relatively small with around 2,000 frames sampled from only 3 videos. Since frames are selected by simply sampling at a fixed rate, many frames are similar to each other in appearance. In addition to datasets with per-pixel labels, Shan et al. \cite{shan2020understanding} labeled 100K video frames with bounding box labels for hands and interacting objects. More recently, Shan et al. \cite{shan2021cohesiv} also proposed to learn hand and hand-held objects segmentation from motion using image and hand location as inputs. 

Our work differs from previous works in two main aspects. While previous work mainly focus on egocentric hand segmentation, we take a step further to study not only hand segmentation but also interacting object segmentation. In addition, previous datasets were mainly focused on certain constrained scenes and limited activities. Our proposed dataset includes diverse daily activities with hundreds of human subjects. More detailed comparisons are shown in Section~\ref{sec:dataset}.

\subsection{Hand-Object Interaction}

Many works have studied hand-object interaction from different angles other than segmentation. One highly related direction is to model and estimate 3D hand joints \cite{cai2018weakly,moon2018v2v,mueller2018ganerated,romero2010hands,sharp2015accurate,sridhar2013interactive,tagliasacchi2015robust,tzionas2016capturing,yang2019disentangling,ye2016spatial,yuan2018depth,zimmermann2017learning,brahmbhatt2020contactpose} and mesh \cite{kulon2020weakly,kulon2019single,pavlakos2019expressive,rong2020frankmocap,xiang2019monocular,zhou2020monocular,muller2021self}, object pose \cite{gkioxari2019mesh,kundu20183d,kuo2020mask2cad,lim2013parsing,michel2017global,sahasrabudhe2019lifting,sun2018pix3d,tulsiani2018factoring,xiang2017posecnn,zhang2020perceiving}, or both \cite{cao2021reconstructing,hasson2020leveraging,hasson2021towards,hasson2019learning,tekin2019h+}. A line of works \cite{brahmbhatt2019contactgrasp,corona2020ganhand,jiang2021hand,karunratanakul2020grasping,taheri2020grab} have also attempted to generate hand pose conditioned on objects. Mostly related to our work, Hasson et al. \cite{hasson2021towards} and Cao et al. \cite{cao2021reconstructing} used segmentation masks of hands and interacting objects to compute 2D projection loss in order to optimize the 3D mesh reconstructions of hand-object pairs. However, the instance segmentation model \cite{kirillov2020pointrend} they used to pre-compute hand and object masks are pretrained on COCO \cite{lin2014microsoft}, which is not tailored to egocentric hand-object segmentation, and thus heavy human intervention is often needed to fix or filter out wrongly predicted masks. Other directions of hand-object interaction involve using hands as probes for object understanding \cite{goyal2021human}, affordance hotspot reasoning \cite{nagarajan2019grounded,nagarajan2020ego}, or even leveraging visual hand-object understanding for robotic learning \cite{mandikal2021dexvip,mandikallearning,nagarajan2021shaping}. Overall, we view our work as an orthogonal foundational tool for many of these vision tasks. 

% \newpage

\section{Dataset}
\label{sec:dataset}

% \subsection{Gathering Data from Multiple Sources}

\textbf{Gathering Data from Multiple Sources.} A big motivation of this work is that the existing datasets do not support researchers to train a model that generalizes well in the wild. Therefore, we collect data from multiple sources, including 7,458 frames from Ego4d \cite{grauman2021ego4d}, 2,121 frames from EPIC-KITCHENS \cite{damen2018scaling}, 806 frames from THU-READ \cite{tang2017action}, as well as 350 frames of our own collected indoor egocentric videos. This results in a total of 11,243 frames sparsely sampled from nearly 1,000 videos covering a wide range of daily activities in diverse scenarios. We manually select diverse and non-repetitive video frames from the sampled set that contain interesting hand-object 
interactions to label with per-pixel segments, as shown in Fig.~\ref{fig:label_demo}.
More details on video frame sampling are included in the supplementary materials.

% \input{figs/data_sources}

% To extract diverse frames from these large-scale video datasets, we first sparsely extract frames for every a few seconds depending on the video sources. For example, in the Epic Kitchen \cite{damen2018scaling} dataset, we extract one frame at every three seconds from episodes across all the subjects, which result in 220K+ frames. Since we only want to label frames containing hand or hand-object interaction, we train a binary classifier to distinguish whether an image contains hand or not for all the candidate frames. The classifier is trained on 4,348 positive examples (with hand) and 649 negative examples (no hand), and can reach around 95\% test accuracy. When applying the classifier on all the candidate frames, the classifier finds 190,426 frames with hand and 31,884 frames without hand. Finally, we evenly sampled 2,121 frames across different subjects from the filtered candidate frames for Epic Kitchen \cite{damen2018scaling}. We apply the similar strategy to filter and sample frames for Ego4d \cite{grauman2021ego4d} dataset, and manually select interesting frames to able for the other two datasets. As a result, we select 2,121 frames from Epic Kitchen, 6,xxx frames from Ego4d \cite{grauman2021ego4d}, 806 frames from THU-READ \cite{tang2017action}, and 350 frames from our own egocentric videos. Overall, this sums up to a 10,xxx frames. 

\input{figs/label_demo}

% \subsection{Annotations for Hands and Interacting Objects Segmentation}

\textbf{Annotations. } For every image in the dataset, we obtained the following per-pixel mask annotations if applicable: (a) left-hand; (b) right-hand; (c) left-hand object; (d) right-hand object; (e) two-hand object. For each type of interacting object, we also provide two levels of interaction: direct and indirect interaction. We define direct interaction between hand and object if the hand touches the objects, such as the blue, cyan, or pink masks in Fig.\ref{fig:label_demo}. Otherwise, we label the object as indirectly interacted with by the hand if the object is being indirectly interacted with, without touching, such as the light cyan masks in the third row of Fig.\ref{fig:label_demo}. In this work, we only study directly interacting objects, but we will release the data to support future research into indirect interacting object segmentation. Note that previous works define hand masks in two types: hand only \cite{bambach2015lending,urooj2018analysis} and hand with arm \cite{li2018eye}. We think both types of labels are useful depending on the application, so we provide both types of hand mask labels for all images in our dataset, where one for hands and another one for the rest of the arms. 

\input{tables/datasets}

\textbf{Comparison with Existing Datasets. } In Table \ref{tab:datasets}, we compare our proposed dataset with existing labeled datasets. 100-DOH \cite{shan2020understanding} also provide a large volume of labelled images and objects, but its labels are at the bounding box level and not tailored towards egocentric images only. Although 100-DOH \cite{shan2020understanding} has made a great effort to improve the generalization of hand-object bounding box detection, we think that having the segmentation prediction is particularly useful or necessary for many downstream vision applications, such as mesh reconstruction of hand-object interaction and seeing through the hands, as shown in Section~\ref{sec:applications}. Compared to other segmentation datasets, one important characteristic of our dataset is that our images cover diverse activities and many human subjects. Since we do not have frame-level semantic labels, our conservative estimation of the number of human subjects and activity types are 300+ and 100+ respectively, according to the video IDs/names in the datasets \cite{damen2018scaling,grauman2021ego4d,tang2017action}. Both the number of subjects and activities are orders of magnitude larger than previous segmentation datasets. In addition, unlike previous segmentation datasets, we are also the first to provide per-pixel mask labels for the interacting objects.

\section{Hand-Object Contact Boundary}
\label{sec:hand_obj_seg}

% The difference between this work and other segmentation task: causal reasoning

A key challenge of hand-object segmentation is the explicit understanding and modeling of the relationship between the hand and the interacting object. Segmenting the object purely based on appearance, as in traditional segmentation tasks, would not properly solve our problem. The reason is that the same object requires segmentation in certain frames but not in the others, depending on whether the hand is in contact with the object.  To this end, we propose to explicitly model the interaction relationship between hand and object by introducing the notion of a dense contact boundary.

% The goal of this work is to robustly segment the left/right hand and the interacting objects by each hand. Traditional instance or semantic segmentation aims to learn a semantic grouping of pixels into multiple classes in an image. Our hand-object segmentation is seemingly similar to previous formulation, but fundamentally different and more challenging in two aspects. First, for the object segmentation, we do not want to segment all the existing objects in the image but only the ones that are interacted by hands. In the words, the same object is needed to be segmented in certain frames but not in the others, depending on the hand-object interaction state. Thus, purely semantic grouping of pixels based appearance would not properly solve this task, and could produce lots of false positive. The second difference is that the output channels in semantic segmentation represent different semantic categories, while, in our case, the output channels indicate whether the object is interacted by left hand, right hand, or both hands. In this situation, again, the same object could goes into any one of the output channels, depending on the hand-object interaction. 

\input{figs/model_pipeline}

Conceptually, the dense contact boundary is defined to be the contact region between the hand and the interacting object. In our implementation, we first dilate both the labeled hand and the object masks in an image, then find the overlapped region between the two dilated masks, and finally binarize the overlapped region as our pseudo-ground truth for contact boundary, as shown in the yellow region in Fig. \ref{fig:model_pipeline}. With such a pipeline, we automatically generate supervision on the contact boundary for all images, where we could train the network to make prediction for it with standard binary cross entropy loss. 

The advantages of explicitly predicting a dense contact boundary for interacting object segmentation are: 1) the contact boundary could provide a cue as to whether there is an interacting object for a given hand mask; 2) it also provides a clearer hand-object separation cue to improve segmentation accuracy. Our experiments show that the contact boundary helps the segmentation model to achieve a higher averaged object mask performance, and more ablation studies are shown in Section~\ref{sec:hand_and_interacting_ob_seg}. Other advantages of the contact boundary besides boosting segmentation performance include:  1) the contact boundary segmentation contains crucial information for many downstream tasks, such as activity recognition and 3D mesh modeling of hand and object; 2) it could also provide potential metrics for evaluating segmentation, specifically for object-hand segmentation during an interaction.

We experiment with one hand-object segmentation pipeline that uses dense interaction boundary as an intermediate stage output. We sequentially predict first the left/right hand, then the contact boundary, and finally the interacting object in three stages, as shown in the left of Fig. \ref{fig:model_pipeline}. In each stage, we concatenate the outputs from previous stages as additional inputs. For example, the left/right hand masks are concatenated with the RGB image as inputs to predict the contact boundary; and in the last stage, the RGB image, hand masks, and contact boundary masks are concatenated as inputs to predict the interacting object masks. Our model is built by sequentially stacking networks, which we tried both a convolutional architecture (ResNet-18 backbone \cite{he2016deep} and HRNet head \cite{wang2020deep}) and a transformer architecture (Swin-L backbone \cite{liu2021swin} and UperNet head  \cite{xiao2018unified}). Note that we do not focus on the architecture and loss design in this work, and more training details are described in the supplementary materials.

\vspace{-5 pt}
\section{Context-Aware Compositional Data Augmentation}
\vspace{-5 pt}

Copying-and-pasting foreground instances at different locations into different background scenes has shown to be a simple and effective data augmentation technique for object detection and instance segmentation, as shown in \cite{fang2019instaboost,ghiasi2021simple,zhang2020learning}. In order to further expand the dataset and improve our model performance, we build a context-aware compositional data augmentation pipeline such that the new composite image has semantically consistent foreground (hand-object) and background context.

\input{figs/compositional_augmentation}

Our overall pipeline design is shown in Fig.~\ref{fig:hand_object_annotations}. In the first step, we need to find the ``clean" background scenes that do not contain any hands or interacting objects. The reason is that the image should only contain one egocentric viewer's hands and interacting objects after the composition. To this end, we propose two ways to generate ``clean" background. The first is to build a simple binary classifier that finds the frames with no hands from a large pool of video frames, as shown in the top left of Fig.~\ref{fig:hand_object_annotations}. The second way is to remove the existing hand-object using an image inpainting model \cite{suvorov2022resolution} and the labeled segmentation masks, as shown in the top right of Fig.~\ref{fig:hand_object_annotations}. Both approaches enable us generate a large pool of ``clean" background candidates. On the other hand, when given an image with hand-object segmentation masks, we first inpaint the hand-object regions using the inpainting model \cite{suvorov2022resolution} to generate the ``clean" query background, and then use it to retrieve the top-K similar background scenes from the ``clean" background candidate pool based on deep features extracted by \cite{simonyan2014very}. Finally, multiple background scenes are sampled from the top-K retrieved background images, as shown in the bottom of Fig.~\ref{fig:hand_object_annotations}. Overall, our designed context-aware image composition pipeline allows us to generate semantically consistent hand-object and context as much as needed. In the experiments, we show the effectiveness of our proposed data augmentation technique.

% \newpage

\section{Experiments on Hand-Object Segmentation}
\label{sec:experiments}

In this section, we first make comparison studies with the existing datasets on hand segmentation, and then we discuss the benchmark performance of the hand-object segmentation with an ablation study. In order to evaluate the in-the-wild segmentation performance, we sparsely sampled 500 frames from 30 collected Youtube egocentric videos to label as our out-of-distribution test set. In the following segmentation experiments, all results are evaluated on this test set unless otherwise specified. All of our models are trained and evaluated using the MMSegmentation codebase\footnote[1]{MMSegmentation github: \href{https://github.com/open-mmlab/mmsegmentation}{https://github.com/open-mmlab/mmsegmentation}}.

% \vspace{-5pt}
\subsection{Two-Hand Segmentation}
% \vspace{-5pt}

Previous hand segmentation datasets have different definitions of hand labels, such as left/right hand \cite{bambach2015lending}, binary hand \cite{urooj2018analysis} or binary hand + arm \cite{li2018eye}. Since our datasets provide all these types of hand labels, we compare individually with the previous datasets in their settings. For a fair comparison, we train the same ResNet-18 backbone \cite{he2016deep} and HRNet head \cite{wang2020deep} on each dataset, select the best checkpoints based on the validation set, and finally compute the results on the same held-out test set. 

\input{tables/leftright_hand_segmentation}

\input{tables/binary_hand_segmentation}

\input{tables/binary_handarm_segmentation}

\input{figs/two_hand_seg}

In the first type of labeling, EgoHand \cite{bambach2015lending} labeled the left and right hands of people in egocentric activities. Similarly, 100-DOH \cite{shan2020understanding} also labeled large-scale left/right hands but with only bounding box annotations. We compare with 100-DOH by training a weakly supervised segmentation model, BoxInst \cite{tian2021boxinst}, which learns to segment objects given bounding box annotation. To make the hand segmentation performance as good as possible for 100-DOH, we pre-trained BoxInst on 2000 frames sampled from EPIC-KITCHEN. As shown in Table~\ref{tab:leftright_hand_segmentation}, the model trained on our dataset significantly outperforms the model trained on EgoHand and 100-DOH with BoxInst. From the visual results, as shown in the first row of Fig. \ref{fig:two_hand_seg}, we observe that models trained on EgoHand and 100-DOH often generate wrong mask categorical labels, which causes significantly lower mIoU (mean Intersection over Union) for left/right hand segmentation. When we binarize the predicted left/right hand masks of EgoHand and 100-DOH and evaluate them on the binary hand segmentation task, the performance gap bridges closer to us, as shown in Table \ref{tab:binary_hand_segmentation}. This again shows that mis-classification of left/right hand is indeed a major issue that causes low mIoU in Table \ref{tab:leftright_hand_segmentation} for the models trained on \cite{bambach2015lending,shan2020understanding}.

The other two datasets EYTH \cite{urooj2018analysis} and EGTEA \cite{li2018eye} provide only binary mask labels for both hands without differentiating between left or right. EYTH \cite{urooj2018analysis} labeled only the hand region, and EGTEA \cite{li2018eye} labeled both hand and arm regions. In Table \ref{tab:binary_hand_segmentation} and Table \ref{tab:binary_handarm_segmentation}, the quantitative results show that the model trained on our datasets also outperforms previous datasets by an obvious margin in both ``hand" and ``hand + arm" settings, and visual comparisons are shown in the bottom of Fig. \ref{fig:two_hand_seg}. In all these hand segmentation settings, we observe that our context-aware compositional data augmentation (CCDA) consistently improves the hand segmentation performance quantitatively. More qualitative comparisons are included in the supplementary materials. 

% \newpage

% \vspace{-5pt}
\subsection{Hand and Interacting Object Segmentation}
% \vspace{-5pt}
\label{sec:hand_and_interacting_ob_seg}

Since our dataset is the first to provide mask labels for interacting objects, we discuss the benchmark performance of hand-object segmentation with an ablation study in this section. In this task, we assign hands to left and right categories, and objects to three categories based on the interacting hand: left-hand object, right-hand object, and two-hand object. A naive solution is to train a segmentation network that decodes five channels of outputs in parallel, as shown in the $1^{st}$ row of Table \ref{tab:handobj_segmentation}. However, this might not be ideal, since parallel decoding of outputs does not leverage any explicit understanding of the hand-object relationship, as discussed in Section \ref{sec:hand_obj_seg}. Thus, we propose to try to sequentially decode the hand first, and then use predicted left/right hand mask information to explicitly guide the interacting object segmentation, as shown in the $3^{rd}$ row of Table \ref{tab:handobj_segmentation}. We also studied adding contact boundary (CB) as intermediate guide information, and found that it effectively boosts the object segmentation performance, as shown in the comparison between $3^{rd}$ and $5^{th}$ rows. More details about contact boundary are discussed in Section \ref{sec:hand_obj_seg}. Finally, we evaluated the effectiveness of our context-aware compositional data augmentation (CCDA) by integrating it on top of both parallel and sequential models. As shown in the comparison between rows $1^{st}, 3^{rd}, 5^{th}$ and rows $2^{nd}, 4^{th}, 6^{th}$, CCDA slightly improves the left/right hand segmentation and significantly boosts the object segmentation performance. We think the reasons are that compositional augmentation enables the network to learn the pixel grouping of objects more easily when placing them into many different background. More details are on how we choose the quantity of augmented images are are discussed in the supplementary materials. The qualitative results for dense contact boundary prediction and hand-object segmentation in diverse activities are shown in Fig. \ref{fig:contact_boundary} and Fig. \ref{fig:hand_object_seg}, respectively.

\input{tables/handobj_segmentation}

\input{figs/contact_boundary}

\input{figs/hand_object_seg}

\section{Applications}
\label{sec:applications}

In the previous section, we have shown that our hand-object segmentation model achieves reliable performance both quantitatively and qualitatively in diverse scenarios. In this section, we demonstrate that our hand-object segmentation system can be applied to boost or enable several useful and interesting vision applications.

% \vspace{-5pt}
\subsection{Boosting Hand State Classification and Activity Recognition}

Understanding the hand state and recognizing types of activities in egocentric videos are important for human behavior analysis. Similarly to 100-DOH \cite{shan2020understanding}, we define hand states for both left and right hands as the following: (contact with) portable, (contact with) stationary, no-contact, self-contact, and not-exist. The goal of this task is to classify a correct state for each of the two hands of the egocentric viewer, where we use two classification heads to handle this. To this end, we labeled the hand states of 3,531 frames from EPIC-KITCHEN \cite{damen2018scaling} dataset with diverse hand-object interaction. During training, we adopt 8:1:1 ratio to split train, val, and test sets. As shown in Table \ref{tab:hand_state_classification}, by adding hand mask and hand-object masks into the input channel, a classifier with the same backbone \cite{simonyan2014very} could effectively improve its classification performance compared to the baseline that uses RGB images only. For the video activity recognition, we used a subset data of EPIC-KITCHEN Action Recognition benchmark \cite{damen2018scaling} as well as its evaluation protocol. With the SlowFast network \cite{feichtenhofer2019slowfast}, we show that by adding hand masks into training, the top-1 classification accuracy of ``verbs/nouns" boosts from 23.95\%/36.77\% to 25.98\%/37.04\%. A visual illustration of these two tasks is shown in Fig. \ref{fig:hs_ar}.

\input{tables/hand_state_classification}

\input{figs/hs_ar}

\subsection{Improved 3D Mesh Reconstruction of Hand-Object Interaction}

Mesh reconstruction of hand-object interaction is a useful but very challenging task. One mainstream approach \cite{cao2021reconstructing,hasson2021towards} to solve this task is to jointly optimize the 3D scale, translation and rotation of a given 3D object model, as well as the MANO parameters \cite{romero2017mano} for hand. Such optimization process often relies on the estimated hand and object segmentation masks to compute the 3D mesh to 2D projection error. In previous works, researchers \cite{cao2021reconstructing,hasson2021towards} leverage the 100-DOH's \cite{shan2020understanding} detector to localize the hand and interacting object at bounding box level, and then use PointRend \cite{kirillov2020pointrend} pre-trained on COCO \cite{lin2014microsoft} to segment the hand and object masks. The interacting object mask is assigned by a heuristic that the object mask with highest confidence score is the one in interaction. 

\input{figs/hand_mesh}

In this work, we integrate our robust hand-object segmentation model into the previous mesh reconstruction pipeline \cite{hasson2021towards}. Since our hand-object segmentation could generalize better than the previous segmentation component, we enable the hand-object reconstruction generalize in more diverse scenarios with higher visual fidelity. As shown in the first row of Fig. \ref{fig:hand_mesh}, the previous segmentation pipeline oftentimes fails to segment the complete object, and thus the object was optimized into a wrong 3D pose, while our accurate hand-object segmentation enables the object mesh reconstruction to be more accurate. In the second row of Fig. \ref{fig:hand_mesh}, we observe that the previous segmentation pipeline sometimes completely misses the interacting object at the bounding box detection stage, and thus no segmentation and 3D mesh could be generated. In contrast, our pipeline provides higher recall on the object detection, and thus is able to recover the object mesh, as shown in the bottom right of Fig. \ref{fig:hand_mesh}.

% failure cases: 1). mask quality not good; 2). failed to detect the object bounding box completely

\subsection{Seeing Through the Hand in Egocentric Videos}

Finally, in this work, we propose a new interesting application, where the goal is to see through the hand in egocentric videos. With our robust per-frame segmentation of hand masks, we use the recent flow-guided video inpainting algorithm \cite{gao2020flow} to completely remove the hands such that we could see the original content occluded by hands in the videos. A visual example of this application is shown in Fig. \ref{fig:see_thru_hand}, where the hand is removed and the bottles and fridge layers that are originally occluded can now been visualized in every video frame. More video results are included in the supplementary materials. In the egocentric videos, since hands are prevalent and almost moving all the time, they create large occlusions of visual contents. The practical use of our ``hand see through" system is that we could potentially enable the vision system analyze more previously occluded information, for example, in the future AR system.

\input{figs/see_thru_hand}

\section{Conclusion} 
\label{sec:conclusion}

We created a fine-grained egocentric hand-object segmentation dataset and synthetic data augmentation method to 1) enable robustness against out-of-distribution domain change and 2) support downstream tasks.  Our labeled dataset of 11,243 images contains both per-pixel segmentation labels of hand and interacting objects and dense contact boundaries.  Our context-aware compositional data augmentation technique significantly improves segmentation performance, especially for interacting objects. We show that our robust hand-object segmentation model can serve as a foundational tool for several vision applications, including hand state classification, activity recognition, 3D mesh reconstruction of hand-object interaction, and seeing through the hand in egocentric videos. 

%In this work, we propose a dataset of 11,243 images with per-pixel segmentation labels of hand and interacting objects. Our proposed dataset enable us train a robust hand segmentation model that generalizes well to the out-of-the-distribution videos, significantly outperforming the previous datasets. In order to boost the hand-object segmentation, we introduce dense contact boundary to explicitly model hand-object relationship, which is proven to be effective. We also propose a context-aware compositional data augmentation technique, which also significantly improve the segmentation performance, especially for the interacting objects. Finally, we show that our robust hand-object segmentation model can be served as a foundation tool for several vision applications, including hand state classification, egocentric video activity recognition, 3D mesh reconstruction of hand-object interaction, and seeing through the hand in egocentric videos. 

\textbf{Acknowledgment.}
This research is based on work supported by Toyota Research Institute and Adobe Gift Funding. The views and conclusions contained herein are those of the authors and should not be interpreted as representing the official policies, either expressed or implied, of the sponsors.

\input{supplemental}

\clearpage
\bibliographystyle{splncs04}
\bibliography{egbib}
\end{document}

%% file: figs/overview.tex
\begin{figure*}[!t]
    \centering
    %[trim=left bottom right top,
    \includegraphics[trim=0.0in 4.5in 1.2in 0in, clip,width=\textwidth]{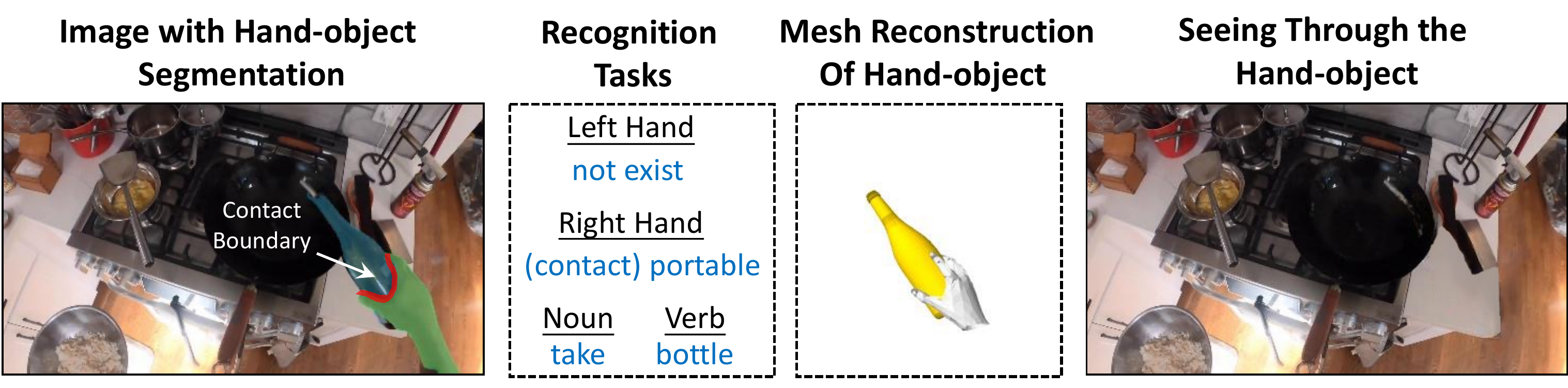}
    \vspace{-20 pt}
    \caption{ Leftmost image: our proposed dataset enables us to train a robust hand-object segmentation model. We introduce contact boundaries to model the hand-object interaction explicitly. Right: our hand-object segmentation model is helpful for many vision tasks, including recognizing hand state, activities, mesh reconstruction, and seeing-through the hand-object. }%An overview of our work.
    \label{fig:overview}
    \vspace{-15 pt}
\end{figure*}

%% file: figs/label_demo.tex
\begin{figure*}[h]
    \vspace{-20 pt}
    \centering
    %[trim=left bottom right top,
    \includegraphics[trim=0.0in 1.2in 2.8in 0in, clip,width=\textwidth]{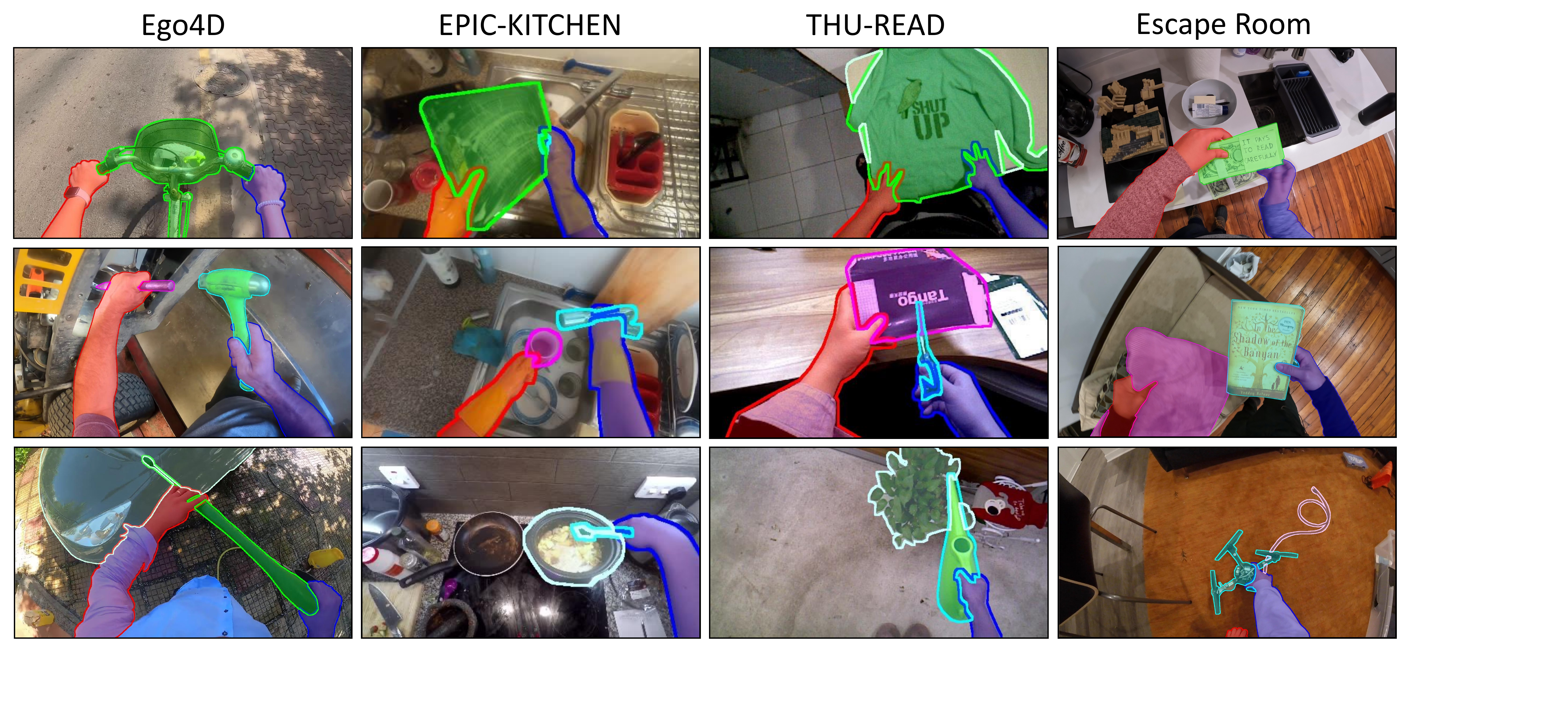}
    \vspace{-20 pt}
    \caption{A selection of images from multiple sources which we label with per-pixel hand and object segments. Color mapping: red $\rightarrow$ left hand, blue $\rightarrow$ right hand, green $\rightarrow$ object interacted by both hands, pink $\rightarrow$ object interacted by left hand, cyan $\rightarrow$ object interacted by right hand. }
    \label{fig:label_demo}
    \vspace{-15 pt}
\end{figure*}

%% file: tables/datasets.tex
\begin{table*}[h!]
   \vspace{-20 pt}
    \begin{center}
     \resizebox{\textwidth}{!}{
      \begin{tabular}
      {lccccccccc}
        \toprule % <-- Toprule here
        \textbf{Datasets} & ~Label~ & ~\#Frames~ & ~\#Hands~ & ~\#Objects~ & ~Objects~ & ~Interaction~ & ~L/R Hand~ & ~\#Subjects~ & ~\#Activities \\
        \midrule % <-- Midrule here
        100-DOH \cite{li2018eye} & box. & 100K & 189.6K & 110.1K & Yes & Yes & Yes & - & -  \\
        \midrule
        EGTEA \cite{li2018eye} & seg. & 13,847 & - & - & Yes & No & No & 32 & 1  \\
        \midrule
        EgoHand\cite{bambach2015lending} & seg.  & 4,800 & 15,053 & - & Yes & - & Yes & 4 & 4  \\
        \midrule
        EYTH \cite{urooj2018analysis} & seg. & 1,290 & 2,600 & - & No  &  No & No & - & -  \\
        \midrule
        Ours & seg. & 11,243 & 20,701 & 17,568 & Yes  & Yes & Yes & 100+ & 300+  \\
        \bottomrule % <-- Bottomrule here
      \end{tabular}
      }
      \vspace{5 pt}
      \caption{\textbf{Egocentric Hand-Object Segmentation Datasets Comparison.} Unknown information is denoted with a dash "-". Compared to previous datasets, our proposed datasets cover relatively diverse scenes and activities with fine-grained segmentation labels of both hands and interating objects. }
      \label{tab:datasets}
    \end{center}
    \vspace{-30 pt}
\end{table*}

%% file: figs/model_pipeline.tex
\begin{figure*}[h]
    \vspace{-10 pt}
    \centering
    %[trim=left bottom right top,
    \includegraphics[trim=0.0in 4.0in 1.0in 0in, clip,width=\textwidth]{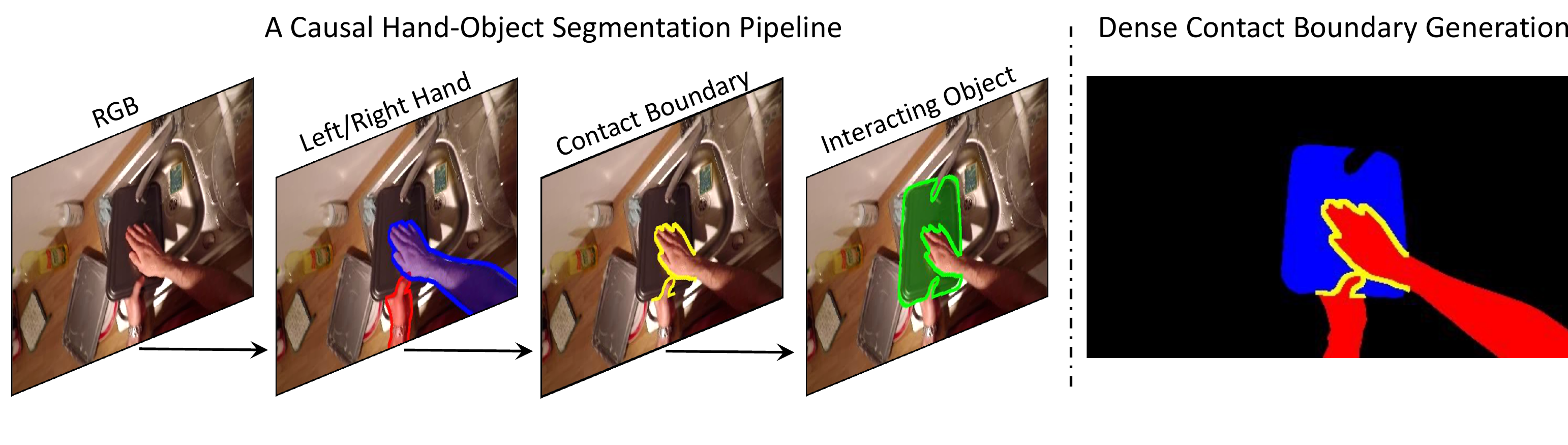}
    \vspace{-20 pt}
    \caption{\textbf{Left}: an overview of our causal hand-object segmentation pipeline. \textbf{Right}: a demo to show how dense contact boundary is defined. }
    \label{fig:model_pipeline}
    \vspace{-15 pt}
\end{figure*}

%% file: figs/compositional_augmentation.tex
\begin{figure*}[h]
    \centering
    \vspace{-10 pt}
    %[trim=left bottom right top,
    \includegraphics[trim=0.2in 0.5in 0.2in 0in, clip,width=\textwidth]{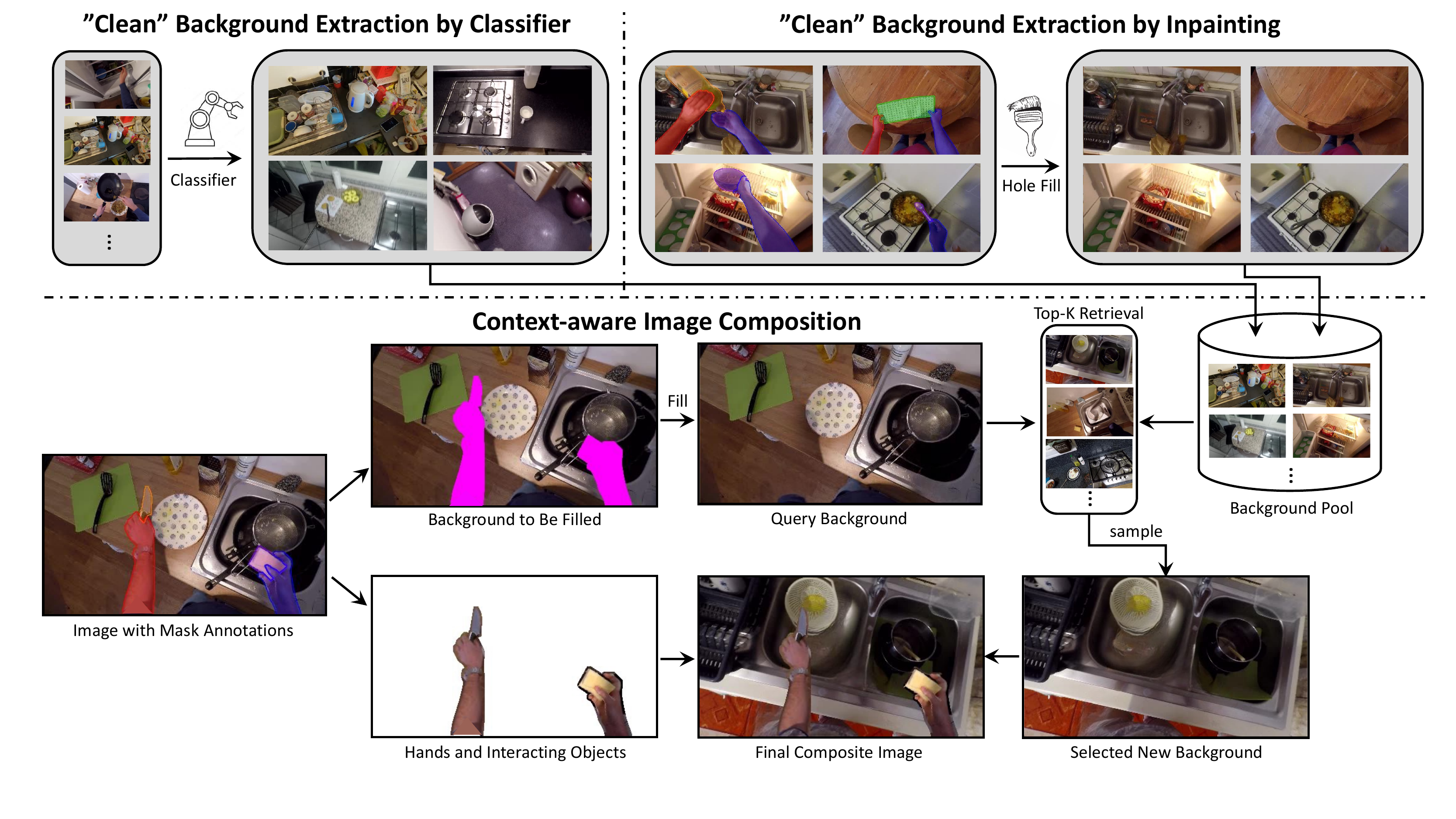}
    \vspace{-20 pt}
    \caption{An overview of our context-aware compositional data augmentation pipeline. }
    \label{fig:hand_object_annotations}
    \vspace{-10 pt}
\end{figure*}

%% file: tables/leftright_hand_segmentation.tex
\begin{table*}[h!]
 \vspace{-15 pt}    
    \begin{center}
     \resizebox{\textwidth}{!}{
      \begin{tabular}
      {lcccc}
        \toprule % <-- Toprule here
        \textbf{Datasets} & \ \hspace{5pt} mIoU \hspace{5pt} & \ \hspace{5pt} mPrec \hspace{5pt} & \ \hspace{5pt} mRec \hspace{5pt} & \ \hspace{5pt} mF1 \hspace{5pt} \\
        \midrule % <-- Midrule here
        EgoHand\cite{bambach2015lending}  & \hspace{5pt} 10.68/33.28 \hspace{5pt} & \hspace{5pt} 43.61/43.20 \hspace{5pt} & \hspace{5pt} 12.39/59.16 \hspace{5pt} & \hspace{5pt} 19.30/49.93 \hspace{5pt} \\
        \midrule
        100-DOH\cite{shan2020understanding}  + BoxInst\cite{tian2021boxinst}  \hspace{5pt} & 36.30/37.51 & 50.06/61.63 & 56.91/48.94 & 53.27/54.55  \\
        \midrule
        Ours & 76.29/77.00 & 83.39/87.06 & 89.97/86.95 & 86.55/87.00  \\
        \midrule
        + CCDA & 79.73/82.17 & 84.26/90.38 & 93.68/90.04 & 88.72/90.21  \\
        \bottomrule % <-- Bottomrule here
      \end{tabular}
      }
      \vspace{2 pt}
      \caption{Left/Right Hand Segmentation. }
      \label{tab:leftright_hand_segmentation}
    \end{center}
    \vspace{-30 pt}
\end{table*}

%% file: tables/binary_hand_segmentation.tex
\begin{table*}[h!]
 \vspace{-5pt}
    % \begin{center}
    \begin{minipage}[t]{0.5\textwidth}
     \centering
      \begin{tabular}
      {p{0.27\textwidth}p{0.16\textwidth}p{0.17\textwidth}p{0.16\textwidth}p{0.16\textwidth}}
        \toprule % <-- Toprule here
        \textbf{Datasets} & \ mIoU & \ mPrec & \ mRec & \ mF1 \\
         \midrule
        EgoHand\cite{bambach2015lending} & 56.51 & 76.33 & 68.52 & 72.22  \\
        \midrule % <-- Midrule here
       
        100-DOH\cite{shan2020understanding}+ BoxInst\cite{tian2021boxinst}& 69.50 & 84.80 & 79.67 & 82.00  \\
        \midrule
         EYTH\cite{urooj2018analysis}  & 75.94 & 85.17 & 87.51 & 86.32  \\
        \midrule
        Ours & 83.18 & 89.34 & 92.34 & 90.82  \\
        \midrule
        + CCDA & 85.45 & 90.11 & 94.3 & 92.15 \\
        \bottomrule % <-- Bottomrule here
      \end{tabular}
      \vspace{5 pt}
      \caption{Binary Hand Segmentation. }
      \label{tab:binary_hand_segmentation}
      \end{minipage}
      \begin{minipage}[t]{0.5\textwidth}
      \centering
      \begin{tabular}
      {p{0.25\textwidth}p{0.16\textwidth}p{0.17\textwidth}p{0.16\textwidth}p{0.16\textwidth}}
        \toprule % <-- Toprule here
        \textbf{Datasets} & \ mIoU & \ mPrec & \ mRec & \ mF1 \\
        \midrule % <-- Midrule here
        % EYTH(#FIXME) & 2.06 & 65.30 & 2.09 & 4.04   \\
        % \midrule
        EGTEA \cite{li2018eye} & 33.26 & 38.24 & 71.87 & 49.92   \\
        \midrule
        Ours & 92.46 & 96.67 & 95.50 & 96.08  \\
        \midrule
        + CCDA & 95.20 & 97.68 & 97.40 & 97.54  \\
        \bottomrule % <-- Bottomrule here
      \end{tabular}
      \vspace{5 pt}
       \caption{Binary Hand + Arm Segmentation. }
      \label{tab:binary_handarm_segmentation}
       \end{minipage}
    %   
    % \end{center}
    \vspace{-10 pt}
\end{table*}

%% file: tables/binary_handarm_segmentation.tex
% \begin{table*}[h!]
%     \begin{center}
%      \resizebox{0.45\textwidth}{!}{
%       \begin{tabular}
%       {l|c|c|c|c}
%         \toprule % <-- Toprule here
%         \textbf{Datasets} & \ mIoU & \ mPrec & \ mRec & \ mF1 \\
%         \midrule % <-- Midrule here
%         % EYTH(#FIXME) & 2.06 & 65.30 & 2.09 & 4.04   \\
%         % \midrule
%         EGTEA & 33.26 & 38.24 & 71.87 & 49.92   \\
%         \midrule
%         Ours w/o CCDA & 92.46 & 96.67 & 95.5 & 96.08  \\
%         \midrule
%         Ours & 95.20 & 97.68 & 97.4 & 97.54  \\
%         \bottomrule % <-- Bottomrule here
%       \end{tabular}
%       }
%       \vspace{5 pt}
%       \caption{Binary Hand + Arm Segmentation. }
%       \label{tab:binary_handarm_segmentation}
%     \end{center}
%     % \vspace{-15 pt}
% \end{table*}

%% file: figs/two_hand_seg.tex
\begin{figure*}[!h]
    \centering
    % \vspace{-10 pt}
    %[trim=left bottom right top,
    \includegraphics[trim=0.0in 2.8in 0.0in 0in, clip,width=\textwidth]{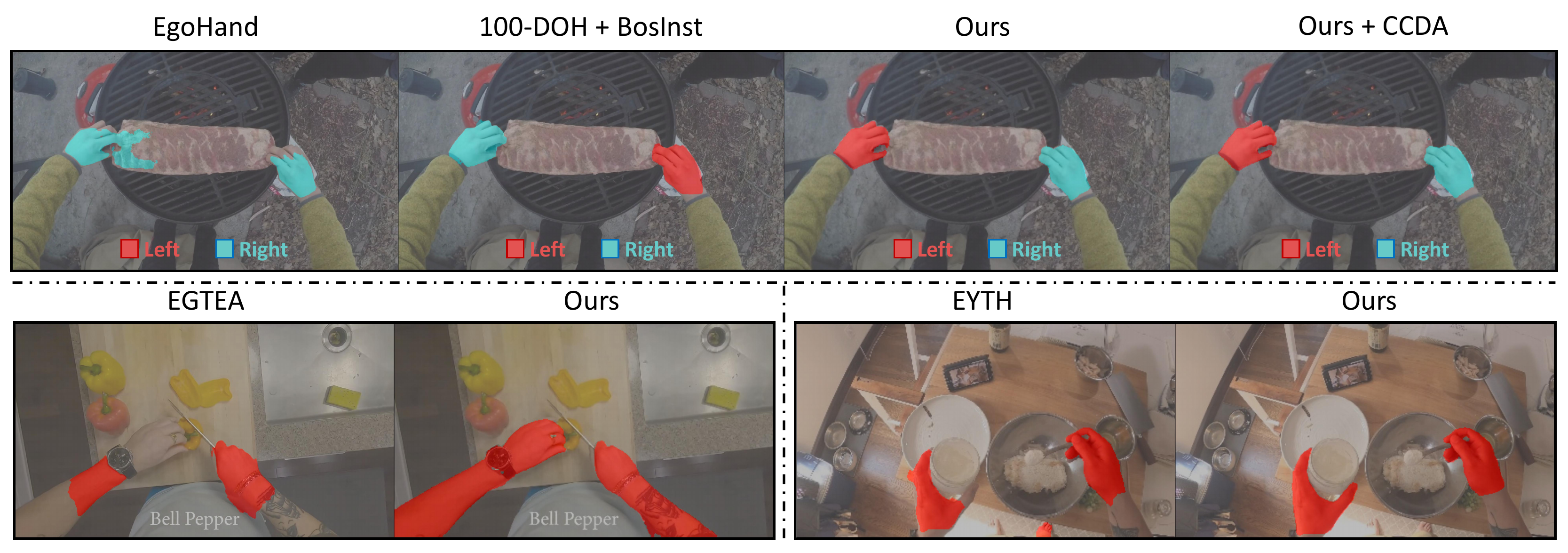}
    \vspace{-20 pt}
    \caption{A qualitative comparison between segmentation models trained on previous datasets and our proposed dataset. The \textbf{top row} shows the comparison with EgoHand and 100-DOH + BoxInst in Left/Right Hand segmentation, where red and cyan indicate left hand and right hand respectively. The \textbf{bottom left} shows the comparison with EGTEA on binary Hand + Arm segmentation. The \textbf{bottom right} shows the comparison with EYTH on binary Hand segmentation.}
    \label{fig:two_hand_seg}
    \vspace{-10 pt}
\end{figure*}

%% file: tables/handobj_segmentation.tex
\begin{table*}[h!]
 \vspace{-10 pt}
    \begin{center}
     \resizebox{\textwidth}{!}{
      \begin{tabular}
      {lccccc}
        \toprule % <-- Toprule here
        \textbf{Models} & \ Left Hand & \ Right Hand & \ Left-Hand Object & \ Right-Hand Object & \ Two-Hand Object \\
        \midrule % <-- Midrule here
        Para. Decode & 69.08 & 73.50 & 48.67 & 36.21 & 37.46   \\
        \midrule
         Para. Decode + CCDA & 77.57 & 81.06 & 54.83 & 38.48 & 39.14   \\
        \midrule
        Seq. Decode & 73.17 & 80.56 & 54.83 & 38.48 & 39.14  \\
        \midrule
        Seq. Decode + CCDA & 87.70 & 88.79 & 58.32 & 40.18 & 46.24  \\
        \midrule
        Seq. Decode + CB & 77.25 & 81.17 & 59.05 & 40.85 & 49.94  \\
        \midrule
        Seq. Decode + CB + CCDA & 87.70 & 88.79 & 62.20 & 44.40 & 52.77  \\
        \bottomrule % <-- Bottomrule here
      \end{tabular}
      }
      \vspace{2 pt}
      \caption{Ablation study on the hand-object segmentation measured by mIoU. }
      \label{tab:handobj_segmentation}
    \end{center}
    \vspace{-29 pt}
\end{table*}

%% file: figs/contact_boundary.tex
\begin{figure*}[!h]
  \vspace{-20 pt}
    \centering
    %[trim=left bottom right top,
    \includegraphics[trim=0.0in 5.6in 0.2in 0.0in, clip,width=\textwidth]{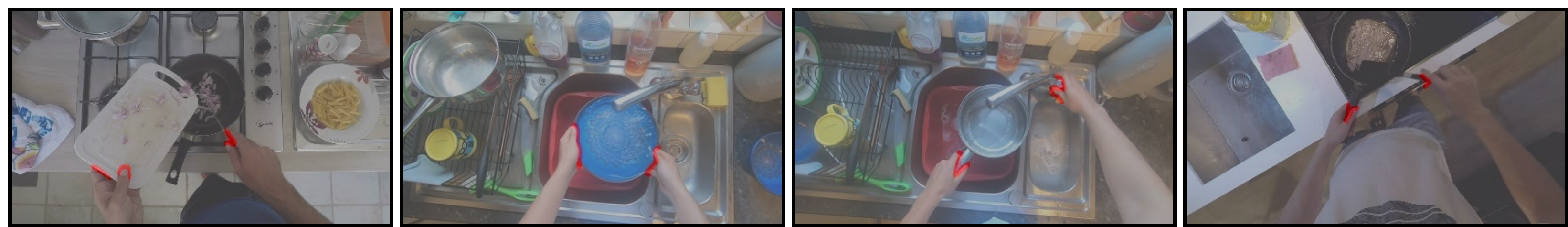}
    \vspace{-15 pt}
    \caption{Qualitative results for the predicted dense contact boundary. }
    \label{fig:contact_boundary}
    % \vspace{-28 pt}
\end{figure*}

%% file: figs/hand_object_seg.tex
\begin{figure*}[!h]
%   \vspace{-10 pt}
    \centering
    %[trim=left bottom right top,
    \includegraphics[trim=0.0in 2.0in 1.4in 0.0in, clip,width=\textwidth]{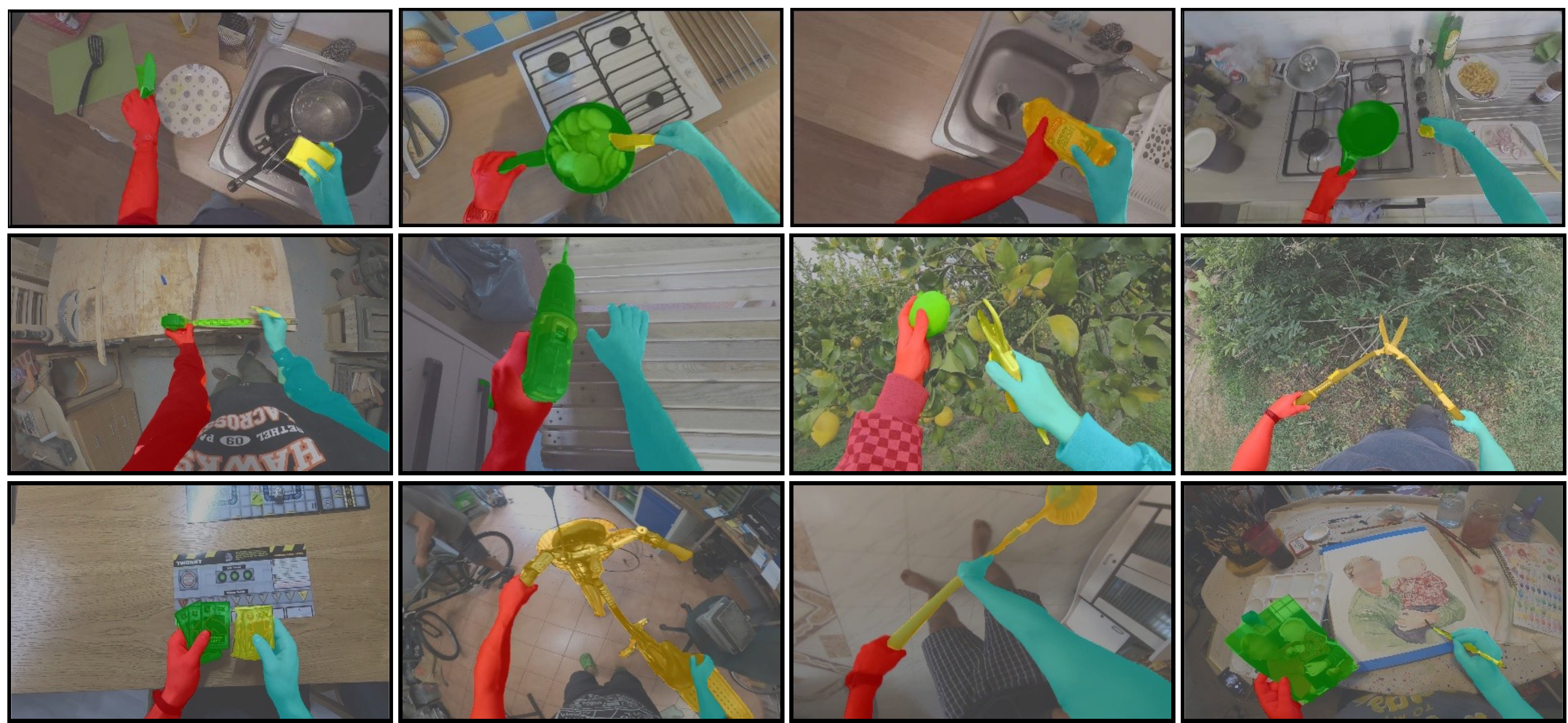}
    \vspace{-20pt}
    \caption{Qualitative results for hand and interating object segmentations on diverse scenarios. The color are coded as follows: red $\rightarrow{}$ left hand, cyan $\rightarrow{}$ right hand, green $\rightarrow{}$ left-hand object, yellow $\rightarrow{}$ right-hand object, orange $\rightarrow{}$ two-hand object. }
    \label{fig:hand_object_seg}
    \vspace{-10 pt}
\end{figure*}

%% file: tables/hand_state_classification.tex
\begin{table*}[h!]
% \vspace{-15 pt}
    \begin{center}
     \resizebox{\textwidth}{!}{
      \begin{tabular}
      {lcccc}
        \toprule % <-- Toprule here
        \textbf{Models} & \ Accuracy & \ Precision & \ Recall & \ F1 score \\
        \midrule % <-- Midrule here
        Baseline 
        & \hspace{5pt} 78.53\%/74.29\% \hspace{5pt} \
        & \hspace{5pt} 43.02\%/37.70\% \hspace{5pt} 
        & \hspace{5pt} 36.86\%/30.60\% \hspace{5pt} 
        & \hspace{5pt} 37.53\%/32.02\% \hspace{5pt} \\
        \midrule
        + Hand Mask & 84.18\%/83.33\% & 64.33\%/66.16\% & 57.31\%/56.34\% & 59.42\%/59.00\% \\
        \midrule
        + Hand \& Object Mask \hspace{5pt} & 86.72\%/83.33\% & 68.18\%/69.04\% & 57.12\%/56.08\% & 60.35\%/59.64\% \\
        \bottomrule % <-- Bottomrule here
      \end{tabular}
      }
      \vspace{5 pt}
      \caption{Quantitative results for left/right hand state classification. }
      \label{tab:hand_state_classification}
    \end{center}
    \vspace{-10 pt}
\end{table*}

%% file: figs/hs_ar.tex
\begin{figure*}[!h]
    % \vspace{-0 pt}
    \centering
    %[trim=left bottom right top,
    \includegraphics[trim=2.2in 4.1in 0.0in 0in, clip,width=\textwidth]{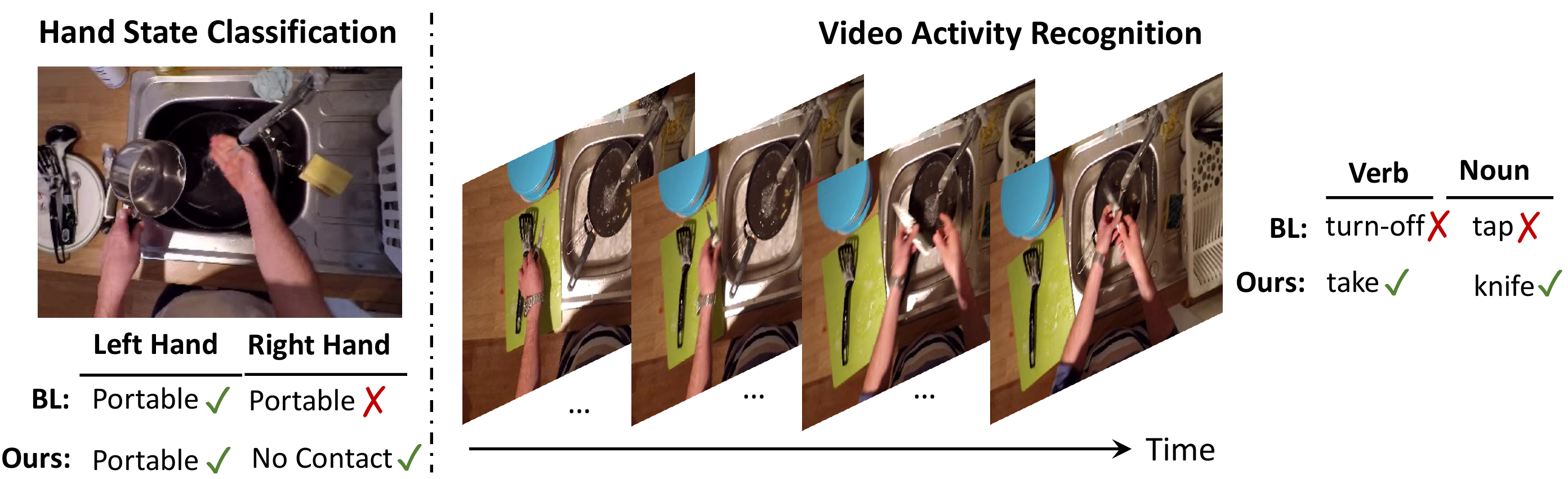}
    % \vspace{-15 pt}
    \caption{A visual comparison example between the baseline and ours for hand state classification and video activity recognition. \textbf{BL} represents the baseline model. }
    \label{fig:hs_ar}
    % \vspace{-10 pt}
\end{figure*}

%% file: figs/hand_mesh.tex
\begin{figure*}[!h]
  \vspace{-20 pt}
    \centering
    %[trim=left bottom right top,
    \includegraphics[trim=0.0in 4.2in 0.1in 0in, clip,width=\textwidth]{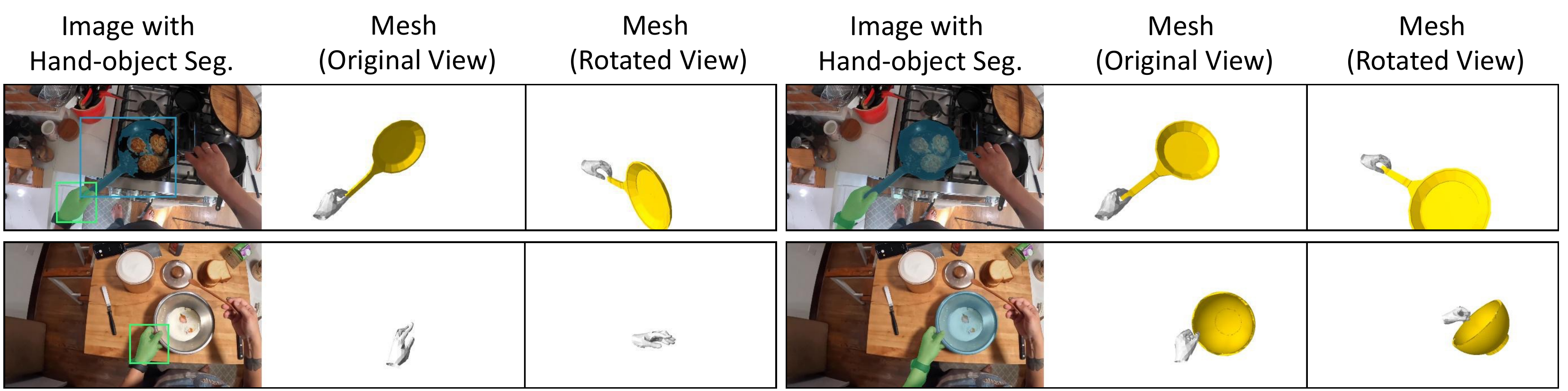}
    \vspace{-15 pt}
    \caption{Visual comparison between 3D mesh reconstruction of hand-object interaction. The \textbf{left} results are from the original code of \cite{hasson2021towards}, where they use 100-DOH \cite{shan2020understanding} detector with PointRend \cite{kirillov2020pointrend} to compute hand-object masks. The \textbf{right} results are computed by integrating our hand-object segmentation into \cite{hasson2021towards} for mesh optimization.  }
    \label{fig:hand_mesh}
    \vspace{-10 pt}
\end{figure*}

%% file: figs/see_thru_hand.tex
\begin{figure*}[!h]
    \vspace{-10 pt}
    \centering
    %[trim=left bottom right top,
    \includegraphics[trim=1.5in 3.6in 2.0in 0in, clip,width=\textwidth]{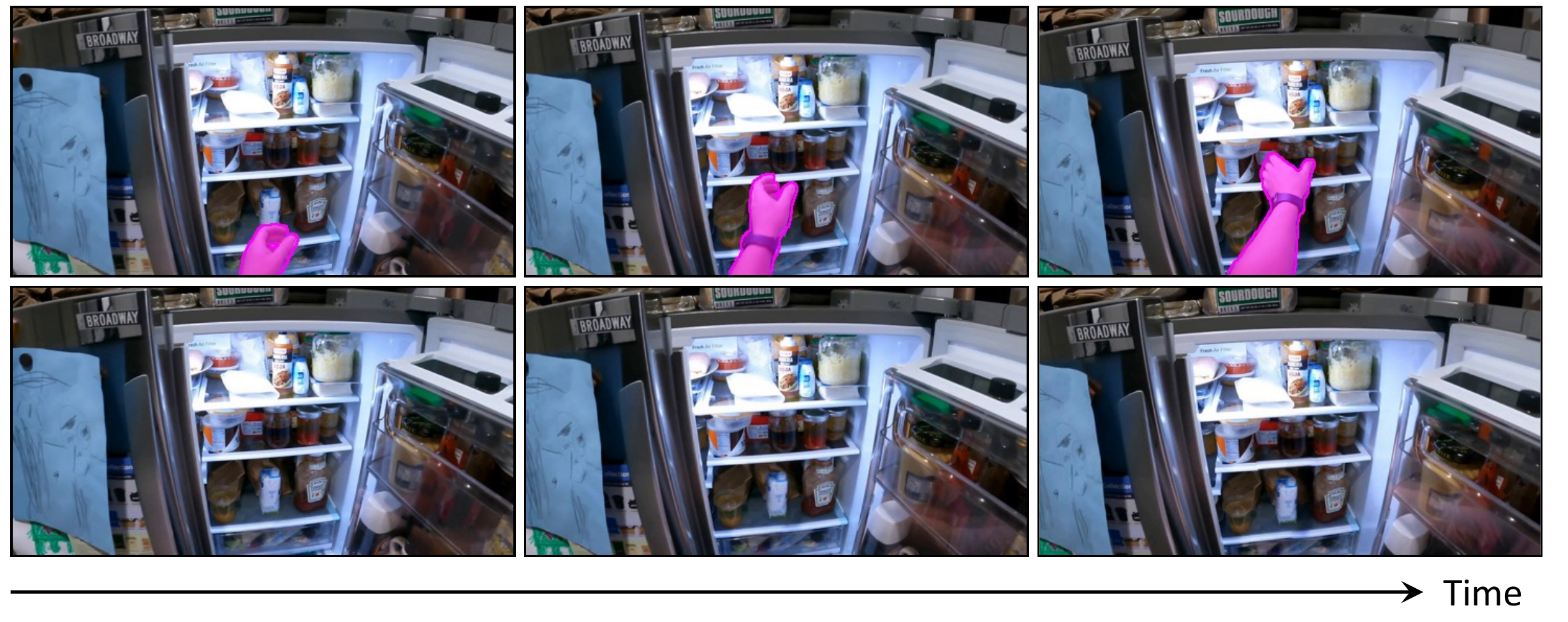}
    \vspace{-20 pt}
    \caption{A qualitative demo to show the application of seeing through the hand in egocentric videos. This application is enabled by our robust per-frame hand segmentation together with the video inpainting model \cite{gao2020flow}. The top row are the frames with predicted hand segmentation masks, and the bottom row shows the ``see through" frames at the corresponding timestamp. More video results are shown in the supplemental.  }
    \label{fig:see_thru_hand}
    \vspace{-10 pt}
\end{figure*}

%% file: supplemental.tex
\title{Supplemental Materials: \\
Fine-Grained Egocentric Hand-Object Segmentation: Dataset, Model, and Applications} % Replace with your title

%\begin{comment}
\titlerunning{ECCV-22 submission ID \ECCVSubNumber} 
\authorrunning{ECCV-22 submission ID \ECCVSubNumber} 
\author{Anonymous ECCV submission}
\institute{Paper ID \ECCVSubNumber}
%\end{comment}
%******************

% \begin{comment}
\titlerunning{Egocentric Hand-Object Segmentation}
\author{Lingzhi Zhang*\inst{1} \and
Shenghao Zhou*\inst{1} \and Simon Stent\inst{2} \and Jianbo Shi\inst{1}}
\authorrunning{Zhang and Zhou et al.}
\institute{University of Pennsylvania \and
Toyota Research Institute \\
}
% \end{comment}
%******************
\maketitle
\footnotetext[1]{* indicates equal contribution}

In this supplementary materials, we first describe the following details of this work: 1). details of video frame collection; 2). training details of segmentation network; 3). how we chose the quantity of data augmentation; 4). the performance of using 100-DOH followed by PointRend. Please note that we also provide results of per-frame hand-object segmentations and "see-through hand" application in egocentric videos. Please check our ".mp4" file in the supplementary materials for details. 

% the details of our dataset formulation, the training details of segmentation network. Then, we analyze how we choose the quantity for our proposed Context-aware Compositional Data augmentation (CCDA). To convert 100-DOH \cite{shan2020understanding} bounding box prediction into segmentation mask, we also compared the performance of using PointRend pretrained on COCO and BoxInst trained on EPIC-KITCHEN. Finally, we show more visual results for our experiments. \textbf{We also include the many results in a separate ".mp4" video file, so please also check our results in the video.}

% \vspace{-10pt}
\section{Dataset}
% \vspace{-5pt}

Our labeled dataset consists of video frames sparsely sampled from multiple sources, including 7,458 frames from Ego4D \cite{grauman2021ego4d}, 2,212 frames from EPIC-KITCHEN \cite{damen2018scaling}, 806 frames from THU-READ \cite{tang2017action}, and 350 frames of our own collected egocentric videos with people playing Escape Room. In Ego4D videos \cite{grauman2021ego4d}, we use the videos from the hand-object interaction challenges, which consist around 1,000 videos. Among these videos, we first sparsely sample one frame per three seconds, and then use 100-DOH detector \cite{shan2020understanding} to filter out the frames that actually contain hand-object interaction. In order to make our labeled data as meaningful as possible, we ask humans to manually select 7,458 frames with diverse and interesting hand-object interactions among these extracted frames. Similarly, we uniformly sample one frame per three seconds in EPIC-KITCHEN \cite{damen2018scaling} videos across 37 participants, then filter out frames that contain hand-object, and finally manually select the interesting frames to label. The THU-READ \cite{tang2017action} dataset consists of short video clips of 40 classes of hand-object interaction, which we uniformly sample a few images in each category. Finally, for our own collected GoPro videos, we also sparsely sample one frame per three seconds, and manually filter out frames to label.

% To extract diverse frames from these large-scale video datasets, we first sparsely extract frames for every a few seconds depending on the video sources. For example, in the EPIC KITCHEN \cite{damen2018scaling} dataset, we extract one frame at every three seconds from episodes across all the subjects, which result in 220K+ frames. Since we only want to label frames containing hand or hand-object interaction, we train a binary classifier to distinguish whether an image contains hand or not for all the candidate frames. The classifier is trained on 4,348 positive examples (with hand) and 649 negative examples (no hand), and can reach around 95\% test accuracy. When applying the classifier on all the candidate frames, the classifier finds 190,426 frames with hand and 31,884 frames without hand. Finally, we evenly sampled 2,121 frames across different subjects from the filtered candidate frames for EPIC KITCHEN \cite{damen2018scaling}. We apply the similar strategy to filter and sample frames for Ego4D \cite{grauman2021ego4d} dataset, and manually select interesting frames to able for the other two datasets. As a result, we select 2,121 frames from EPIC KITCHEN, 6,xxx frames from Ego4D \cite{grauman2021ego4d}, 806 frames from THU-READ \cite{tang2017action}, and 350 frames from our own egocentric videos. Overall, this sums up to a 11,235 frames. 

% \vspace{-10pt}
\section{Training Details of Segmentation Networks}
% \vspace{-5pt}
% Our segmentation network is based on HRNet.
% We use SGD optimizer, with learning rate=0.01, momentum=0.9

In this section, we discuss the details of our segmentation network training. We use ResNet-18 backbone \cite{he2016deep} and HRNet head \cite{wang2020deep} as our base model for all of our experiments. For each experiment, we train the network for 80,000 iterations with SGD optimizer with batch size of 8, learning rate of 0.01, momentum of 0.9, and weight decay of 0.0005. We use random flip and photometric distortion(random brightness, random constrast change, etc.) as our data augmentation techniques on top of our proposed context-aware data augmentation.

% \vspace{-10pt}
\section{Choosing the Augmented Data Quantity}
% \vspace{-5pt}

% compare with randomly sampled background

Our proposed Context-aware Compositional Data Augmentation (CCDA) technique generates the composite data before the training starts, in an offline fashion. A key question is how much data we should generate for the augmented dataset. To this end, we run an experiment to evaluate how the segmentation performance would vary when gradually increasing the number of composite images, as shown in Fig. \ref{fig:CCDA}. We found that the hand and object IoU reaches to the maximum IoU, when augmenting 16K and 8K composite images respectively, on the YouTube testset. 

\input{figs/CCDA}

\section{PointRend vs. BoxInst for 100-DOH}

Since 100-DOH \cite{shan2020understanding} only predicts the bounding of hands and objects, we compare two ways to further convert the bounding boxes to segmentation. In the first way, we use 100-DOH \cite{shan2020understanding} detector to generates pseudo labels of hand-object, and train BoxInst \cite{tian2021boxinst} model to segment hand and objects. In the second way, we use 100-DOH \cite{shan2020understanding} detector to localize the hand-object bounding boxes, and use PointRend \cite{kirillov2020pointrend} to segment the masks. We find that the 100-DOH \cite{shan2020understanding} + PointRend \cite{kirillov2020pointrend} has better performance on the left/right hand segmentation, and 100-DOH \cite{shan2020understanding} + BoxInst \cite{tian2021boxinst} has better performance on the binary hand segmentation, as shown in Table \ref{tab:leftright_hand_segmentation} and \ref{tab:binary_hand_segmentation}, respectively. Nevertheless, the model trained on our dataset surpass the performance both approaches by an obvious margin.

\input{tables/lefthand_hand_segmentation_supp}

\input{tables/binary_hand_segmentation_supp}

We also conduct an experiment to use 100-DOH \cite{shan2020understanding} + PointRend \cite{kirillov2020pointrend} to compute the interacting object segmentation masks. However, we observe that such inference pipeline often completely miss or misclassify the interacting objects segmentation, or in other words has low recall, as shown in Fig. \ref{fig:HOS_PointRend_vs_Ours}. Quantatively, we find that the averaged object IoU of 100-DOH \cite{shan2020understanding} + PointRend \cite{kirillov2020pointrend} is only 12.24, which is significantly lower than ours on the YouTube testset.

%  (36.44)

\input{figs/HOS_PointRend_vs_Ours}

% \section{Compare with 100-DOH on Object Segmentation}

% \input{tables/handobj_segmentation}

% \section{More Visual Results}

% \subsection{Hand Segmentation Comparisons with Other Datasets}

% \subsection{Dense Contact Boundary}

% more visual results for predicted contact boundary.
% Also provide a quantitative benchmark performance. 
% hos comparison when adding cb

% \subsection{Adding CCDA}

% Segmentation comparison when using contact boundary. 

% \subsection{Results for Hand-object Segmentation}

\section{More Segmentation Label Visualization}

We show more visualizations of our hand-object segmentation labels, across different data sources, in Fig .\ref{fig:more_data_labeling}.

\input{figs/more_data_labeling}

\section{More Video Results}

In our project page and github page, we show more results on hand-object segmentation and the application of "seeing through the hand" in the egocentric videos.

%% file: figs/CCDA.tex
\begin{figure*}[!h]
    \centering
    \vspace{-15 pt}
    %[trim=left bottom right top,
    \includegraphics[trim=0.0in 3.3in 2.3in 0.0in, clip,width=\textwidth]{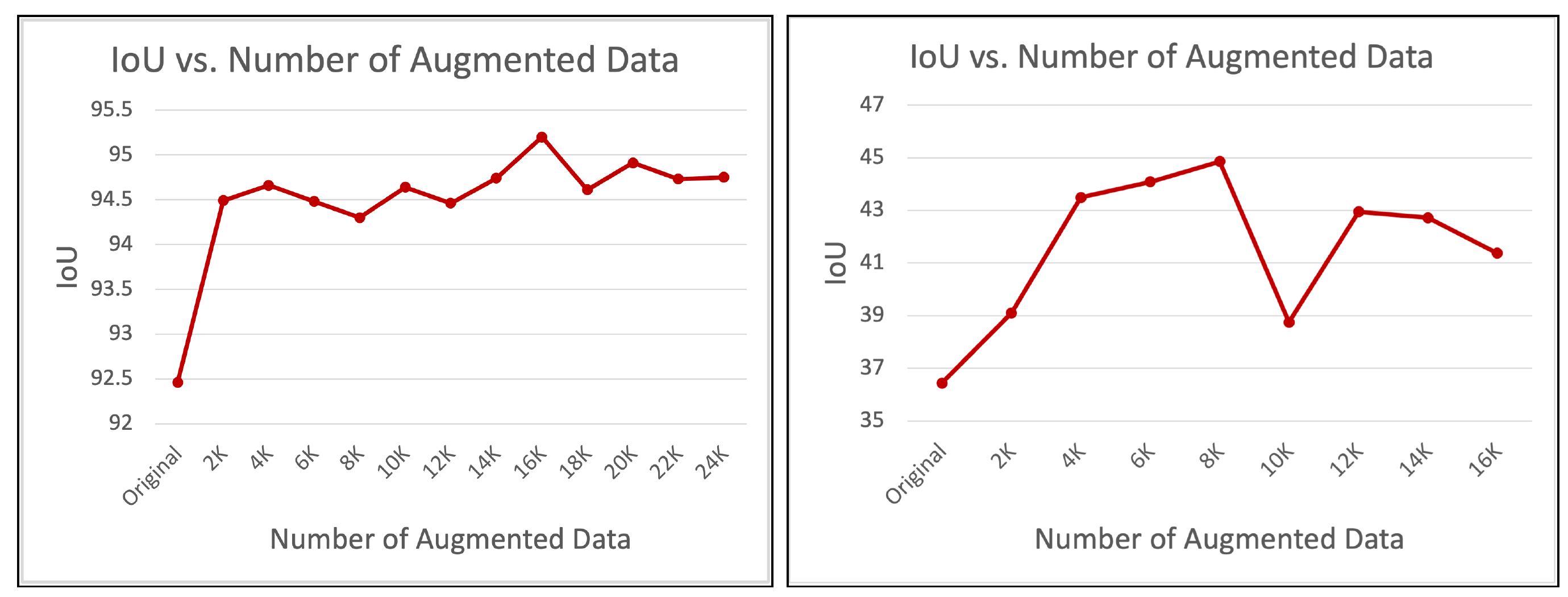}
    \vspace{-20 pt}
    \caption{Averaged hand and object IoU scores vs. the number of augmented data on the left and right, respectively. }
    \label{fig:CCDA}
    \vspace{-30 pt}
\end{figure*}

%% file: tables/lefthand_hand_segmentation_supp.tex
\begin{table*}[h!]
%  \vspace{-15 pt}    
    \begin{center}
     \resizebox{0.95\textwidth}{!}{
      \begin{tabular}
      {l|c|c|c|c}
        \toprule % <-- Toprule here
        \textbf{Datasets} & \ \hspace{10pt} mIoU \hspace{10pt} & \ \hspace{10pt} mPrec \hspace{10pt} & \ \hspace{10pt} mRec \hspace{10pt} & \ \hspace{10pt} mF1 \hspace{10pt} \\
        \midrule % <-- Midrule here
        100-DOH\cite{shan2020understanding}  + BoxInst\cite{tian2021boxinst}  \hspace{5pt} & 36.30/37.51 & 50.06/61.63 & 56.91/48.94 & 53.27/54.55  \\
        \midrule
        100-DOH\cite{shan2020understanding}  + PointRend \cite{kirillov2020pointrend} \hspace{5pt} & 61.83/62.72 & 76.17/78.41 & 76.66/75.8 & 76.41/77.09  \\
        \midrule
        Ours & 79.73/82.17 & 84.26/90.38 & 93.68/90.04 & 88.72/90.21  \\
        \bottomrule % <-- Bottomrule here
      \end{tabular}
      }
       \vspace{5 pt}
      \caption{Left/Right Hand Segmentation. }
      \label{tab:leftright_hand_segmentation}
    \end{center}
    \vspace{-30 pt}
\end{table*}

%% file: tables/binary_hand_segmentation_supp.tex
\begin{table*}[h!]
 \vspace{10 pt}    
    \begin{center}
     \resizebox{0.8\textwidth}{!}{
      \begin{tabular}
      {l|c|c|c|c}
        \toprule % <-- Toprule here
        \textbf{Datasets} & \ \hspace{5pt} mIoU \hspace{5pt} & \ \hspace{5pt} mPrec \hspace{5pt} & \ \hspace{5pt} mRec \hspace{5pt} & \ \hspace{5pt} mF1 \hspace{5pt} \\
        \midrule % <-- Midrule here
        100-DOH\cite{shan2020understanding}  + BoxInst\cite{tian2021boxinst}  \hspace{5pt} & 69.50 & 84.80 & 79.67 & 82.00  \\
        \midrule
        100-DOH\cite{shan2020understanding}  + PointRend \cite{kirillov2020pointrend} \hspace{5pt} & 63.62 & 78.39 & 77.14 & 77.76  \\
        \midrule
        Ours & 85.45 & 90.11 & 94.30 & 92.15 \\
        \bottomrule % <-- Bottomrule here
      \end{tabular}
      }
       \vspace{5 pt}
      \caption{Binary Hand Segmentation. }
      \label{tab:binary_hand_segmentation}
    \end{center}
    \vspace{-30 pt}
\end{table*}

%% file: figs/HOS_PointRend_vs_Ours.tex
\begin{figure*}[!h]
    \centering
    \vspace{-15 pt}
    %[trim=left bottom right top,
    \includegraphics[trim=0.0in 2.8in 0.6in 0.0in, clip,width=\textwidth]{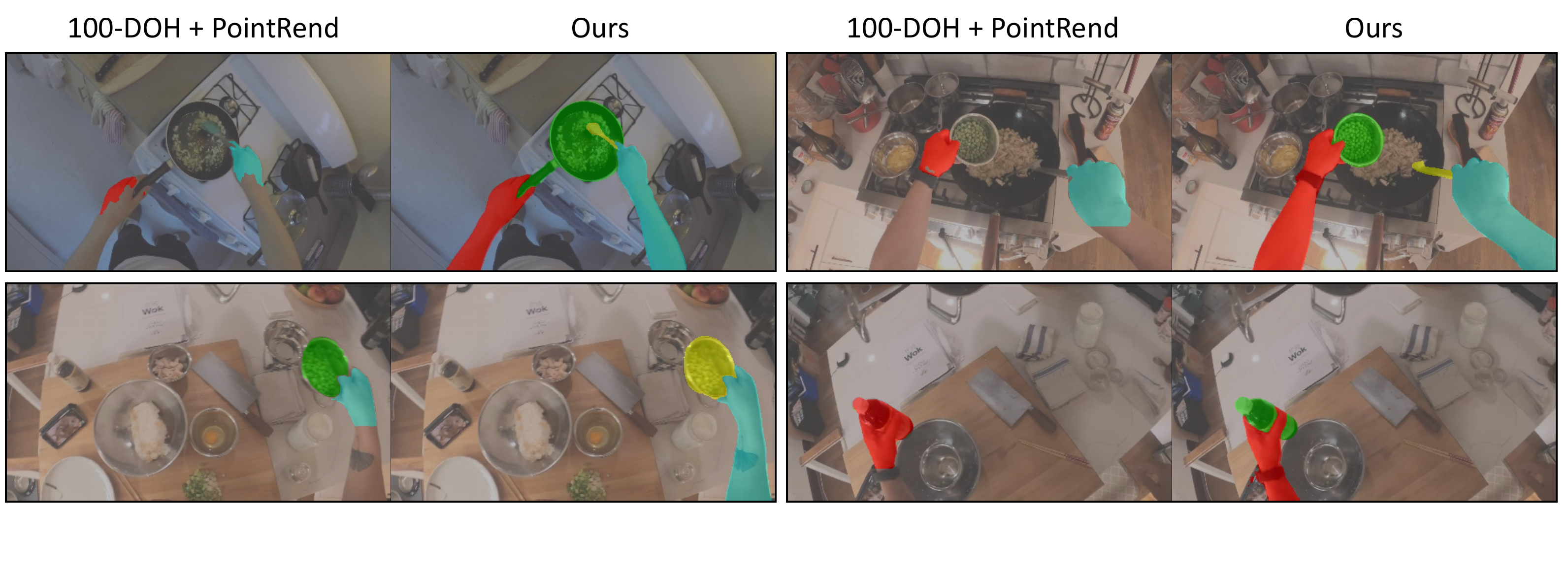}
    \vspace{-30 pt}
    \caption{Visual comparison of hand-object segmentaiton between 100-DOH \cite{shan2020understanding} + PointRend \cite{kirillov2020pointrend} and ours. The color are coded as follows: red $\rightarrow{}$ left hand, cyan $\rightarrow{}$ right hand, green $\rightarrow{}$ left-hand object, yellow $\rightarrow{}$ right-hand object, orange $\rightarrow{}$ two-hand object. }
    \label{fig:HOS_PointRend_vs_Ours}
    \vspace{-20 pt}
\end{figure*}

%% file: figs/more_data_labeling.tex
\begin{figure*}[!t]
    \centering
    \vspace{-15 pt}
    %[trim=left bottom right top,
    \includegraphics[trim=0.0in 3.0in 0.4in 0.0in, clip,width=\textwidth]{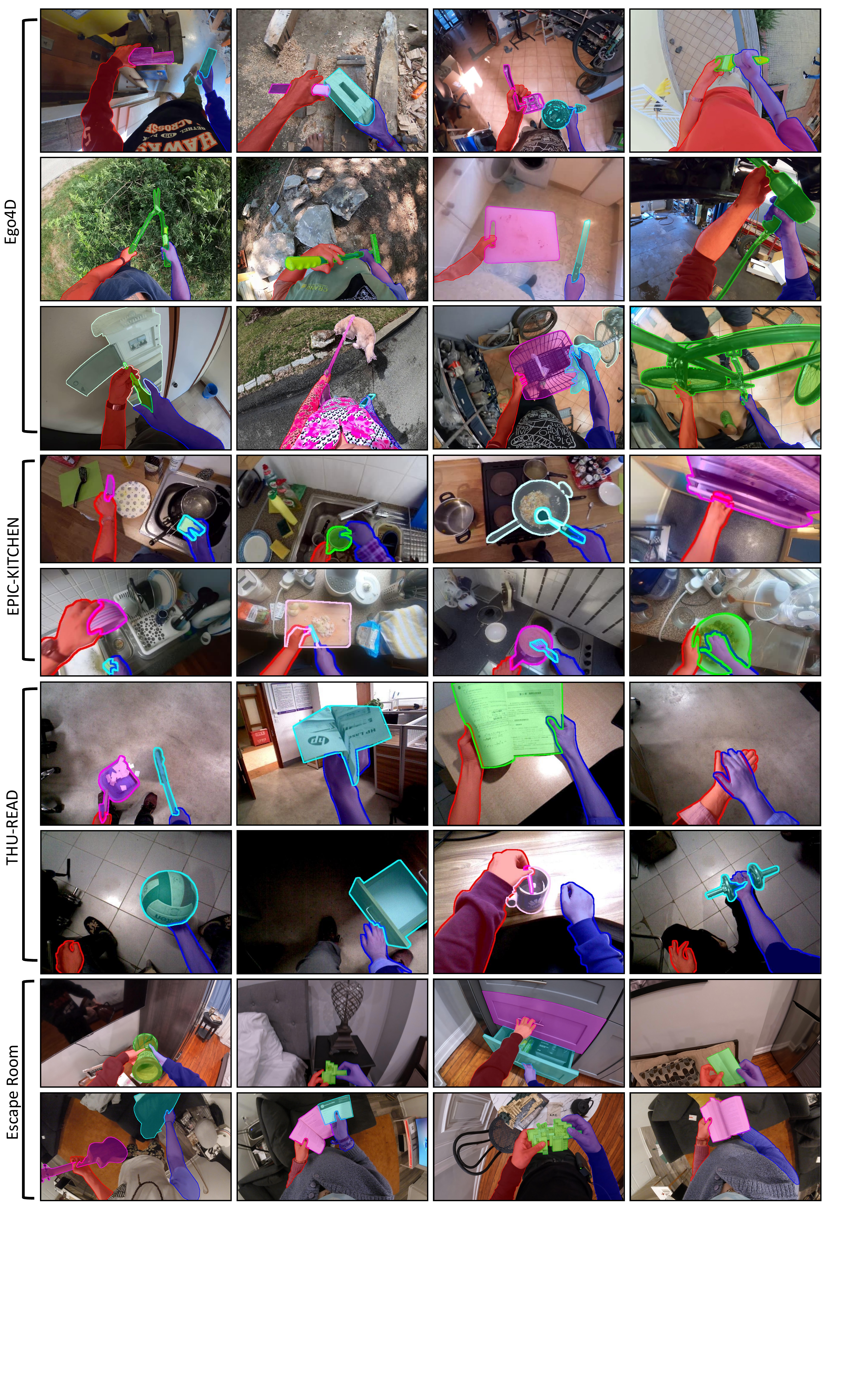}
    \vspace{-20 pt}
    \caption{More visual demonstration of our hand-object segmentation labels. }
    \label{fig:more_data_labeling}
    \vspace{-30 pt}
\end{figure*}